\def\year{2021}\relax
\documentclass[letterpaper]{article} 
\usepackage{aaai21}  
\usepackage{times}  
\usepackage{helvet} 
\usepackage{courier}  
\usepackage[hyphens]{url}  
\usepackage{graphicx} 
\usepackage{natbib}  
\usepackage{caption} 
\usepackage{lscape} 

\usepackage{booktabs}       
\usepackage{nicefrac}       
\usepackage{microtype}      
\usepackage{caption}
\usepackage{listings}
\usepackage{multirow}
\lstdefinestyle{mystyle}{
        language=python,
        basicstyle=\footnotesize\ttfamily,
        breaklines=true,
        keepspaces=true,
    }
\usepackage{adjustbox}
\usepackage{sidecap}
\usepackage{subcaption}
\usepackage{makecell}
\usepackage{amsmath}
\usepackage{lineno}
\usepackage{enumitem}
\usepackage{comment}
\usepackage{array}
\newcolumntype{L}{>{\centering\arraybackslash}m{3cm}}
\usepackage{adjustbox}

\title{Improving Clinical Document Understanding\\ on COVID-19 Research with Spark NLP}
\author{
    Veysel Kocaman,
    David Talby
    \\
}
\affiliations{
    John Snow Labs Inc.\\

    16192 Coastal Highway\\
    Lewes, DE , USA 19958\\
    \{veysel, david\}@johnsnowlabs.com
    
}

\begin{document}

\maketitle

\begin{abstract}
Following the global COVID-19 pandemic, the number of scientific papers studying the virus has grown massively, leading to increased interest in automated literate review. We present a clinical text mining system that improves on previous efforts in three ways. First, it can recognize over 100 different entity types including social determinants of health, anatomy, risk factors, and adverse events in addition to other commonly used clinical and biomedical entities. Second, the text processing pipeline includes assertion status detection, to distinguish between clinical facts that are present, absent, conditional, or about someone other than the patient. Third, the deep learning models used are more accurate than previously available, leveraging an integrated pipeline of state-of-the-art pre-trained named entity recognition models, and improving on the previous best performing benchmarks for assertion status detection. We illustrate extracting trends and insights - e.g. most frequent disorders and symptoms, and most common vital signs and EKG findings – from the COVID-19 Open Research Dataset (CORD-19). The system is built using the Spark NLP library which natively supports scaling to use distributed clusters, leveraging GPU’s, configurable and reusable NLP pipelines, healthcare-specific embeddings, and the ability to train models to support new entity types or human languages with no code changes.

\end{abstract}

\section{Introduction}
\label{sec:introduction}

The COVID-19 pandemic brought a surge of academic research about the virus - resulting in 23,634 new publications between January and June of 2020 \cite{Silva2020PublishingVI} and accelerating to 8,800 additions per week from June to November on the COVID-19 Open Research Dataset \cite{wang2020cord}. Such a high volume of publications makes it impossible for researchers to read each publication, resulting in increased interest in applying natural language processing (NLP) and text mining techniques to enable semi-automated literature review \cite{Cheng2020AnOO}.

In parallel, there is a growing need for automated text mining of Electronic health records (EHRs) in order to find clinical indications that new research points to. EHRs are the primary source of information for clinicians tracking the care of their patients. Information fed into these systems may be found in structured fields for which values are inputted electronically (e.g. laboratory test orders or results)~\cite{liede2015validation} but most of the time information in these records is unstructured making it largely inaccessible for statistical analysis~\cite{murdoch2013inevitable}. These records include information such as the reason for administering drugs, previous disorders of the patient or the outcome of past treatments, and they are the largest source of empirical data in biomedical research, allowing for major scientific findings in highly relevant disorders such as cancer and Alzheimer’s disease ~\cite{perera2014factors}.

\begin{figure*}[htb!]
\includegraphics[width=0.9\textwidth,scale=0.9]{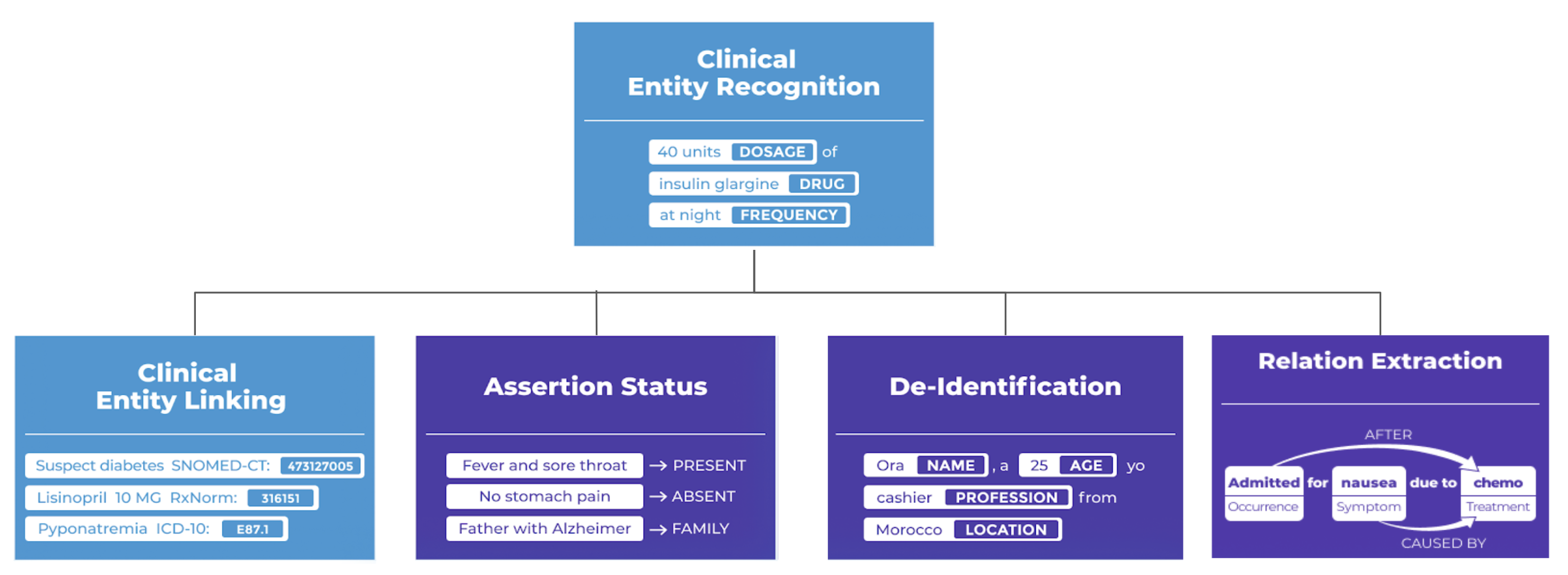}
\centering
\caption{Named Entity Recognition is a fundamental building block of medical text mining pipelines, and feeds downstream tasks such as assertion status, entity linking, de-identification, and relation extraction.}
\label{fig:ner_tree_chart}
\end{figure*}

A primary building block in such text mining systems is named entity recognition (NER) - which is regarded as a critical precursor for question answering, topic modelling, information retrieval, etc~\cite{yadav2019survey}.  In the medical domain, NER recognizes the first meaningful chunks out of a clinical note, which are then fed down the processing pipeline as an input to subsequent downstream tasks such as clinical assertion status detection~\cite{uzuner20112010}, clinical entity resolution ~\cite{tzitzivacos2007international} and de-identification of sensitive data~\cite{uzuner2007evaluating} (see Figure~\ref{fig:ner_tree_chart}). However, segmentation of clinical and drug entities is considered to be a difficult task in biomedical NER systems because of complex orthographic structures of named entities ~\cite{liu2015effects}.

The next step following an NER model in the clinical NLP pipeline is to assign an assertion status to each named entity given its context. The status of an assertion explains how a named entity (e.g. clinical finding, procedure, lab result) pertains to the patient by assigning a label such as present ("patient is diabetic"), absent ("patient denies nausea"), conditional ("dyspnea while climbing stairs"), or associated with someone else ("family history of depression"). In the context of COVID-19, applying an accurate assertion status detection is crucial, since most patients will be tested for and asked about the same set of symptoms and comorbidities - so limiting a text mining pipeline to recognizing medical terms without context is not useful in practice.

In this study, we introduce a set of pre-trained NER models that are all trained on biomedical and clinical datasets within a Bi-LSTM-CNN-Char deep learning architecture, and a Bi-LSTM based assertion detection module built on top of the Spark NLP software library. We then illustrate how to extract knowledge and relevant information from unstructured electronic health records (EHR) and COVID-19 Open Research Dataset (CORD-19) by combining these models in a pipeline. Using state-of-the-art deep learning architectures, Spark NLP's NER and Assertion modules can also be extended to other spoken languages with zero code changes and can scale up in Spark clusters. Moreover, by utilizing Apache Spark, both training and inference of full NLP pipelines can scale to make the most of distributed Spark clusters. Due to brevity concerns, the implementation details and training metrics of these models will be kept out of the scope of this study.

The specific novel contributions of this paper are:
\begin{itemize}

\item Introducing a medical text mining pipeline composed of state-of-the-art, healthcare-specific NER models

\item Introducing a clinical assertion status detection model that establishes a new state-of-the-art level of accuracy on a widely used benchmark

\item Describing how to apply these models in a unified, performant, and scalable pipeline on documents from the CORD-19 dataset.

\end{itemize}

The remainder of the paper is organized as follows: 
Section~\ref{sec:NerDL} Introduces the Spark NLP library, summarizes the NER and assertion detection model frameworks it implements, and elaborates the named entities in each pre-trained NER model.
Section~\ref{sec:parsed_outputs} explains how to build a prediction pipeline to extract named entities and assign assertion statuses from a set of documents on a cluster with Spark NLP.
Section~\ref{sec:scaling} discusses benchmarking speed and scalability issues and
Section~\ref{sec:conclusion}  concludes this paper by summarizing key points and future directions.


\section{Named Entity Recognition in Spark NLP}
\label{sec:NerDL}

\begin{figure}[!htbp]
\includegraphics[width=0.4\textwidth,scale=0.4]{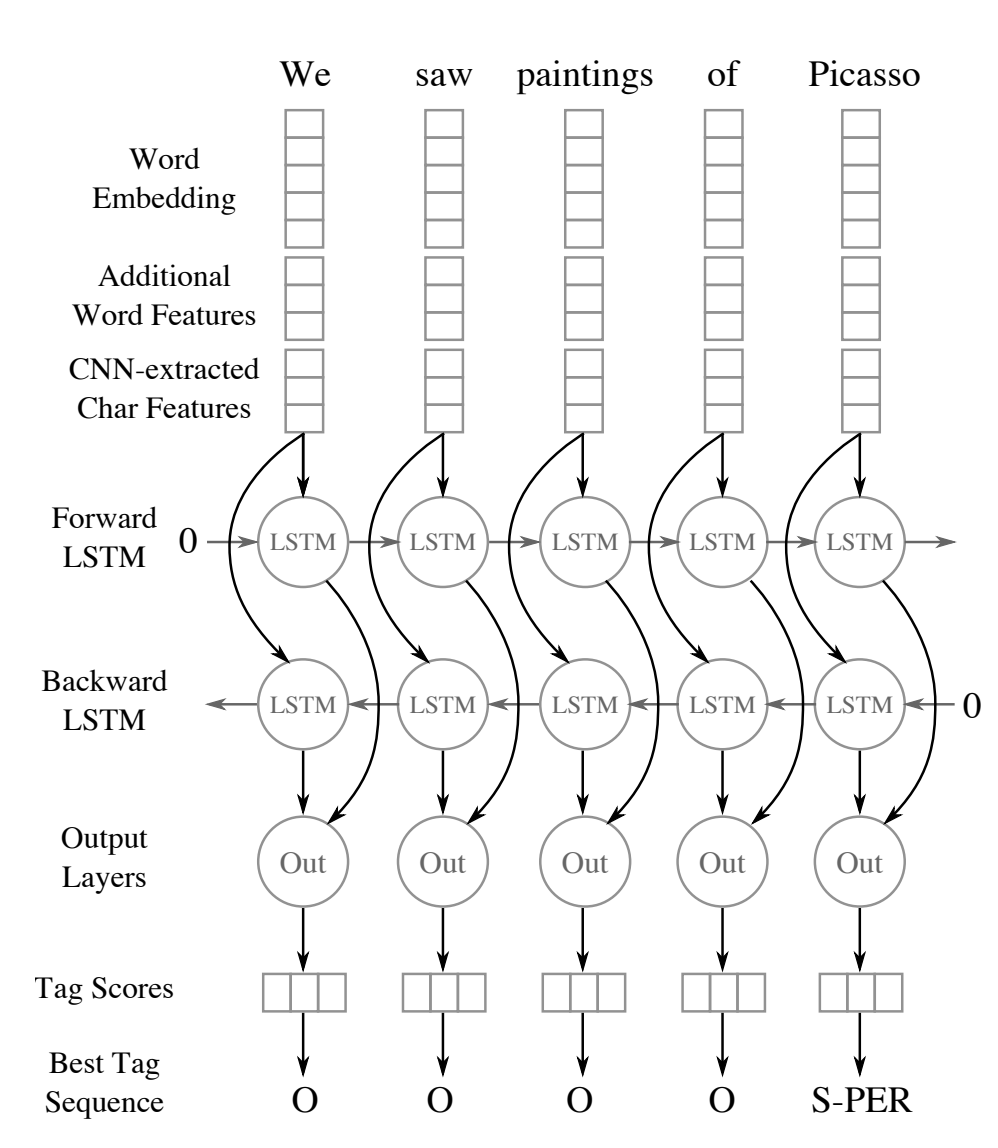}
\centering
\caption{Overview of the original BiLSTM-CNN-Char architecture~\cite{chiu2016named}.}
\label{fig:ner_dl_diagram}
\end{figure}

\begin{figure*}[htb!]
\centering
\includegraphics[width=0.9\textwidth]{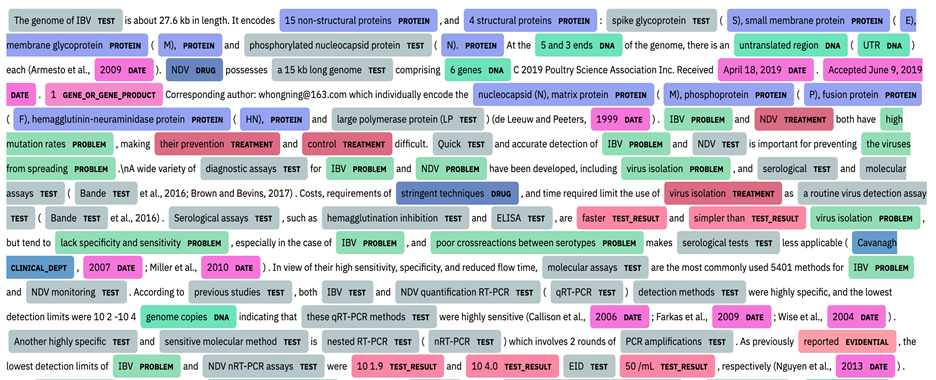}
\centering
\caption{Sample predictions from pre-trained clinical NER models in Spark NLP for Healthcare}
\label{fig:clinical_ner}
\end{figure*}

The deep neural network architecture for named entity recognition in Spark NLP is based on the BiLSTM-CNN-Char framework. It is a modified version of the architecture proposed by Chiu et.al.~\cite{chiu2016named}. It is a neural network architecture that automatically detects word and character-level features using a hybrid bidirectional LSTM and CNN architecture, eliminating the need for most feature engineering steps. The detailed architecture of the framework in the original paper is illustrated at Figure~\ref{fig:ner_dl_diagram} and a sample predictions from a set of pre-trained clinical NER models from a text taken from CORD-19 dataset is shown in ~\ref{fig:clinical_ner}.

In Spark NLP, this architecture is implemented using TensorFlow, and has been heavily optimized for accuracy, speed, scalability, and memory utilization. This setup has been tightly integrated with Apache Spark to let the driver node run the entire training using all the available cores on the driver node. There is a CuDA version of each TensorFlow component to enable training models on GPU when available. The Spark NLP provides open-source API's in Python, Java, Scala, and R - so that users do not need to be aware of the underlying implementation details (TensorFlow, Spark, etc.) in order to use it.

The full list of the entities for each pre-trained medical NER model is available in Appendix~\ref{appendix:entities}, the accuracy metrics are given in Table~\ref{tab:ner_metrics} and a sample Python code for training a NER model from scratch is in Appendix~\ref{appendix:python_code_ner}.

\begin{table}[htbp!]
\caption{Validation metrics of the selected NER models trained with clinical word embeddings in Spark NLP. These NER models are trained with the datasets mentioned in the original papers cited (Appendix~\ref{appendix:entities})} 
\label{tab:ner_metrics} 
\resizebox{0.98\columnwidth}{!}{
\begin{tabular}{llll}
\hline
model & \thead{number of \\ entity} & \thead{micro\\ F1} & \thead{macro\\ F1} \\\hline
ner\_anatomy & 10                                   & 0.750                          & 0.851                         \\
ner\_bionlp & 15                                   & 0.638                         & 0.748                         \\
ner\_cellular & 4                                    & 0.792                         & 0.813                         \\
ner\_clinical & 3                                    & 0.872                         & 0.873                         \\
ner\_deid\_sd & 7                                    & 0.896                         & 0.942                         \\
ner\_deid\_enriched & 17                                   & 0.762                         & 0.934                         \\
ner\_diseases & 1                                    & 0.960                          & 0.960                          \\
ner\_drugs & 1                                    & 0.963                         & 0.964                         \\
ner\_events & 10                                   & 0.690                          & 0.801                           \\
jsl\_ner\_wip & 76                                   & 0.842                         & 0.863                         \\
ner\_posology & 6                                    & 0.881                         & 0.922                         \\
ner\_risk\_factors & 8                                    & 0.593                         & 0.728                         \\
ner\_human\_phenotype\_go & 2                                    & 0.904                         & 0.922                         \\
ner\_human\_phenotype\_gene & 2                                    & 0.871                         & 0.876                         \\
ner\_chemprot & 3                                    & 0.785                         & 0.817                         \\
ner\_ade & 2                                    & 0.824                         & 0.852 \\\hline

\end{tabular}
}
\end{table}

\section{Assertion Status Detection in Spark NLP}
\label{sec:AssertionDL}

The deep neural network architecture for assertion status detection in Spark NLP is based on a Bi-LSTM framework, and is a modified version of the architecture proposed by Fancellu et.al.~\cite{fancellu2016neural}. Its goal is to classify the assertions made on given medical concepts as being \textit{present}, \textit{absent}, or \textit{possible} in the patient, \textit{conditionally} present in the patient under certain circumstances, \textit{hypothetically present} in the patient at some future point, and mentioned in the patient report but \textit{associated with someone-else} ~\cite{uzuner20112010}. 

In the proposed implementation, input units depend on the target tokens (a named entity) and the neighboring words that are explicitly encoded as a sequence using word embeddings. Similar to Fancellu et.al.~\cite{fancellu2016neural} we have observed that that 95\% of the scope tokens (neighboring words) fall in a window of 9 tokens to the left and 15 to the right of the target tokens in the same dataset. We therefore implemented the same window size and used learning rate 0.0012, dropout 0.05, batch size 64 and a maximum sentence length 250. The model has been implemented within Spark NLP as an annotator called \textit{AssertionDLModel}. After training 20 epoch and measuring accuracy on the official test set, this implementation exceeds the latest state-of-the-art accuracy benchmarks as summarized as Table~\ref{tab:assertion_metrics}

\begin{table}[htb!]
\caption{Assertion detection model test metrics. Our implementation exceeds the benchmarks in the latest best model~\cite{uzuner20112010} in 4 out of 6 assertion labels - and in overall accuracy.}
\centering
\label{tab:assertion_metrics}
\begin{tabular}{lll}
\hline
\thead{Assertion\\ Label} & \thead{Spark\\ NLP} & \thead{Latest\\ Best} \\\hline
Absent & 0.944 & 0.937 \\
Someone-else & 0.904 & 0.869 \\
Conditional & 0.441 & 0.422 \\
Hypothetical & 0.862 & 0.890 \\
Possible & 0.680 & 0.630 \\
Present & 0.953 & 0.957 \\\hline
micro F1 & 0.939 & 0.934\\
\hline
\end{tabular}
\end{table}

A sample predictions from a clinical assertion detection model can be seen at Table~\ref{tab:assertiondl}.

\begin{figure*}[ht!]
\includegraphics[width=1.0\textwidth, scale=0.7]{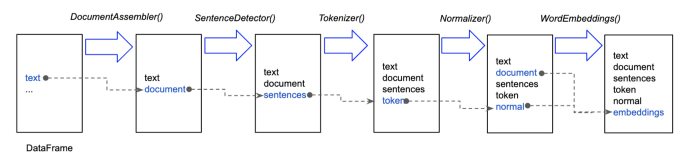}
\centering
\caption{The flow diagram of a Spark NLP pipeline. When we fit() on the pipeline with a Spark data frame, its text column is fed into the DocumentAssembler() transformer and a new column \textit{document} is created as an initial entry point to Spark NLP for any Spark data frame. Then, its document column is fed into the SentenceDetector() module to split the text into an array of sentences and a new column “sentences” is created. Then, the “sentences” column is fed into Tokenizer(), each sentence is tokenized, and a new column “token” is created. Then, Tokens are normalized (basic text cleaning) and word embeddings are generated for each. Now data is ready to be fed into NER models and then to the assertion model. }
\label{fig:pipeline_diagram}
\end{figure*}

\begin{table}[ht!]
\caption{Sample predictions from the pre-trained clinical assertion detection model in Spark NLP. \\ 
\\Sample text : \textit{Patient with severe fever and sore throat. He shows no stomach pain and is maintained on an epidural and PCA for pain control. He also became short of breath with climbing a flight of stairs. After CT, lung tumor located at the right lower lobe. Father with Alzheimer.}}
\centering
\label{tab:assertiondl}
\begin{tabular}{lll}
\hline
chunk & entity & assertion \\\hline
severe fever & PROBLEM & Present \\
sore throat & PROBLEM & Present \\
stomach pain & PROBLEM & Absent \\
an epidural & TREATMENT & Present \\
PCA & TREATMENT & Present \\
pain control & PROBLEM & Present \\
short of breath & PROBLEM & Conditional \\
CT & TEST & Present \\
Lung tumor & PROBLEM & Present \\
Alzheimer & PROBLEM & Someone-else\\\hline
\end{tabular}
\end{table}

\pagebreak

\section{Analysing the CORD-19 Dataset with Pre-trained Models}
\label{sec:parsed_outputs}

Since assertion status labels are assigned to a medical concept that is given as an input to the assertion detection model, NER and assertion models must work together sequentially. In Spark NLP, we handle this interaction by feeding the output of NER models to an NER converter to create chunks from labeled entities and then feed these chunks to the assertion status detection model within the same pipeline. The flow diagram of such a pipeline can be seen in Figure~\ref{fig:pipeline_diagram}. As the flow diagram shows, in Spark NLP each generated (output) column is pointed to the next module as an input, depending on its input column specifications. A sample Python code for such a prediction pipeline can be seen at Appendix~\ref{appendix:python_code_pipeline}.

This enables users to easily configure arbitrary pipelines - such as running 20 NER pre-trained models within one pipeline, as we do in this analysis of the CORD-19 dataset. NLP pipelines configured this way are easily reproducible, since they are seriablizable and directly expressed in code. They also simplify  experimentation - for example, comparing multiple NER and assertion status models in the same run (while benefiting from the fact that data and embeddings are only loaded into memory once), or trying with different text cleaning steps before the NER stage (such as stopword removal, lemmatization, or automated spell correction).

While the CORD-19 text mining pipeline scales to process an arbitrary number of articles, for purposes of concrete demonstration the next two tables show results on a randomly sampled of 100 articles. The number of recognized named entities for the selected entity classes can be seen at Table~\ref{tab:entitiy_stats}.  The number of entities detected from each document (20 NER models, over 10 document) can be seen at Table~\ref{tab:model_doc_stats}. The most frequent phrases from the selected entity types can be found at Table~\ref{tab:topterms}. The predictions from the assertion status detection model for \textit{Disease\_Syndrome\_Disorder} is shown in Table~\ref{tab:cord_ner_assertions}.

\def\rot{\rotatebox}

\begin{table*}[htb!]
\caption{The number of entities for the selected entity classes per document from COR-19 dataset (10 documents sampled).}
\label{tab:entitiy_stats}
\resizebox{0.98\textwidth}{!}{
\begin{tabular}{lllllllllllllllllllll}
\rot{90}{document\_id} & \rot{90}{anatomy} & \rot{90}{cell} & \rot{90}{organism} & \rot{90}{ade} & \rot{90}{dna} & \rot{90}{protein} & \rot{90}{problem} & \rot{90}{treatment} & \rot{90}{test} & \rot{90}{location} & \rot{90}{disease} & \rot{90}{drug} & \rot{90}{drug\_ingredient} & \rot{90}{test\_result} & \rot{90}{medical\_device} & \rot{90}{virus} & \rot{90}{chemical} & \rot{90}{gene} & \rot{90}{chem} & \rot{90}{species} \\
\midrule
1  & 157     & 189  & 280      & 64               & 134 & 150     & 1312    & 944       & 634  & 188      & 124     & 608  & 129              & 30           & 55              & 229   & 109      & 130  & 56   & 254     \\
2  & 277     & 296  & 137      & 120              & 155 & 124     & 1024    & 475       & 620  & 62       & 39      & 243  & 51               & 122          & 76              & 4     & 95       & 188  & 55   & 130     \\
3  & 210     & 252  & 54       & 105              & 33  & 129     & 406     & 388       & 377  & 66       & 52      & 304  & 99               & 39           & 31              & 2     & 94       & 104  & 90   & 26      \\
4  & 94      & 196  & 77       & 76               & 71  & 77      & 479     & 490       & 565  & 31       & 26      & 293  & 71               & 70           & 51              & 70    & 139      & 136  & 97   & 67      \\
5  & 12      & 0    & 14       & 51               & 4   & 3       & 240     & 127       & 145  & 73       & 67      & 89   & 23               & 12           & 23              & 1     & 94       & 42   & 44   & 11      \\
6  & 6       & 7    & 9        & 7                & 8   & 14      & 222     & 90        & 56   & 183      & 54      & 36   & 0                & 2            & 1               & 18    & 15       & 61   & 11   & 44      \\
7  & 29      & 15   & 69       & 54               & 2   & 25      & 384     & 680       & 271  & 29       & 38      & 451  & 76               & 16           & 239             & 99    & 114      & 61   & 53   & 33      \\
8  & 27      & 16   & 25       & 29               & 18  & 420     & 318     & 246       & 443  & 47       & 18      & 165  & 24               & 43           & 5               & 15    & 116      & 134  & 78   & 25      \\
9  & 44      & 17   & 42       & 14               & 0   & 2       & 456     & 93        & 138  & 41       & 169     & 71   & 23               & 7            & 1               & 0     & 14       & 13   & 19   & 18      \\
10 & 1       & 0    & 15       & 1                & 0   & 1       & 42      & 23        & 11   & 22       & 16      & 9    & 0                & 0            & 1               & 0     & 0        & 2    & 0    & 9\\
\hline
\end{tabular}
}
\end{table*}

\begin{table*}[htb!]
\caption{The total number of entities from the selected NER models per document from COR-19 dataset (10 documents sampled).}
\label{tab:model_doc_stats} 
\resizebox{0.98\textwidth}{!}{
\begin{tabular}{lllllllllllllllllllll}
\rot{90}{document\_id} & \rot{90}{anatomy\_coarse} & \rot{90}{anatomy} & \rot{90}{bionlp} & \rot{90}{cellular}& \rot{90}{clinical}& \rot{90}{deid}& \rot{90}{enriched}& \rot{90}{diseases} & \rot{90}{drugs} & \rot{90}{events\_clinical}& \rot{90}{jsl\_ner\_wip}& \rot{90}{medmentions}& \rot{90}{posology}& \rot{90}{risk\_factors} & \rot{90}{human\_phenotype\_go} & \rot{90}{human\_phenotype\_gene}& \rot{90}{chemprot\_clinical} & \rot{90}{ade\_clinical} & \rot{90}{chemicals} & \rot{90}{bacterial\_species} \\
\hline
1  & 157 & 128 & 584 & 336 & 1313 & 487 & 387 & 124 & 62 & 1649 & 847 & 1904 & 144 & 182 & 77  & 129 & 81  & 435 & 56  & 254 \\
2  & 277 & 259 & 713 & 368 & 948  & 184 & 167 & 39  & 51 & 1182 & 772 & 1429 & 101 & 75  & 71  & 125 & 150 & 139 & 55  & 130 \\
3  & 210 & 200 & 511 & 226 & 510  & 130 & 61  & 52  & 90 & 697  & 633 & 924  & 165 & 33  & 75  & 84  & 148 & 178 & 90  & 26  \\
4  & 94  & 93  & 318 & 214 & 656  & 41  & 22  & 26  & 42 & 771  & 525 & 957  & 104 & 18  & 31  & 110 & 98  & 196 & 97  & 67  \\
5  & 12  & 11  & 83  & 7   & 219  & 136 & 88  & 67  & 9  & 425  & 211 & 654  & 21  & 103 & 12  & 49  & 41  & 78  & 44  & 11  \\
6  & 6   & 5   & 31  & 25  & 140  & 215 & 128 & 54  & 0  & 442  & 135 & 590  & 4   & 80  & 75  & 66  & 15  & 35  & 11  & 44  \\
7  & 29  & 37  & 178 & 33  & 637  & 31  & 20  & 38  & 20 & 771  & 404 & 967  & 95  & 15  & 25  & 59  & 59  & 372 & 53  & 33  \\
8  & 27  & 22  & 175 & 441 & 436  & 138 & 46  & 18  & 25 & 579  & 311 & 728  & 69  & 28  & 27  & 102 & 114 & 91  & 78  & 25  \\
9  & 44  & 43  & 106 & 3   & 319  & 66  & 77  & 169 & 19 & 455  & 445 & 478  & 26  & 53  & 24  & 35  & 17  & 43  & 19  & 18  \\
10 & 1   & 4   & 17  & 1   & 35   & 33  & 23  & 16  & 0  & 64   & 49  & 98   & 2   & 19  & 1   & 1   & 1   & 8   & 0   & 9  \\
\hline
\end{tabular}}
\end{table*}

One benefit for this system compared to previous work is the variety of medical entity types that be recognized: As detailed in Appendix~\ref{appendix:entities}, this NLP pipeline extracts over 100 entity types. While most clinical named entity recognition focus on symptoms, treatments, and drugs, and most biomedical focused projects focus on chemicals, proteins, and genes, this pipeline goes beyond these and can also extract:

\begin{itemize}

\item Entities related to social determinants of health such as age and gender, rate and ethnicity, diet, social history, employment, relationship status, alcohol use, sexual activity and orientation

\item Medical risk factors such as hypertension, smoking, cholesterol, hyperlipidemia, weight and BMI, kidney disease, pregnancy, and diabetes

\item Specific vital signs and lab results such as pulse, temperature, O2 saturation,respiration, LDL and HDL

\item Detailed biomedical entity types such as organ, tissue, gene, human phenotype, chrmical, species, amino acid, protein, cell, cell component, biological function, chemical, substance, process

\end{itemize}

Table~\ref{tab:entitiy_stats} shows that this variety is useful in practice in the context of COVID-19 research. On just 10 randomly selected documents and 20 entity types, there are over 60 cases of more than a hundred instances of one entity type found within one paper. Only in fewer than 10\% of the cells there were fewer there 10 entities recognized for a specific entity type in a specific document. This suggests that text mining approaches that ignore these entity types fail to take advantage of a lot of clinical insight that the COVID-19 research papers include.

Table~\ref{tab:cord_ner_assertions} shows how an accurate assertion status detection model can help in filtering this large amount of entities - in order to focus researchers and downstream algorithms on the most clinically relevant insights. In this small sample, 'systemic disease' is a present clinical condition; 'infectious diseases' and 'disorders of immunity' are hypothetical; while 'skin diseases and 'parvovirus' are associated with someone else.

Consider a common use case of building an automated knowledge graph that links patient symptoms to drugs they are taking, existing conditions, or past procedures. The difference between having assertion status detection results, and being able to filter only to symptoms and drugs that positively impact the patient, will have a substantial impact on the accuracy of the bottom-line results. Since more than a thousand entities are recognized in each research paper, and hundreds of thousands of published COVID-19 papers - doing this automatically, accurately, and at scale is required.

\begin{table*}[htb!]
\caption{The most frequent 10 terms from the selected entity types predicted through parsing 100 articles from CORD-19 dataset~\cite{wang2020cord} with an NER model named \textit{jsl\_ner\_wip} in Spark NLP. Getting predictions from the model, we can get some valuable information regarding the most frequent disorders or symptoms mentioned in the papers or the most common vital and EKG findings without reading the paper. According to this table, the most common symptom is \textit{cough} and \textit{inflammation} while the most common drug ingredients mentioned is \textit{oseltamivir} and \textit{antibiotics}. We can also say that \textit{cardiogenic oscillations} and \textit{ventricular fibrillation} are the common observations from EKGs while \textit{fever} and \textit{hyphothermia} are the most common vital signs.}
\label{tab:topterms} 
\resizebox{0.98\textwidth}{!}{
\begin{tabular}{lllllll}
\hline
\thead{Disease Syndrome\\ Disorder} & \thead{Communicable\\ Disease} & Symptom & \thead{Drug\\ Ingredient} & Procedure & \thead{Vital Sign\\ Findings} & \thead{EKG\\ Findings} \\\hline
infectious diseases & HIV & cough & oseltamivir & resuscitation & fever & low VT \\
sepsis & H1N1 & inflammation & biological agents & cardiac surgery & hypothermia & cardiogenic oscillations \\
influenza & tuberculosis & critically ill & VLPs & tracheostomy & hypoxia & significant changes \\
septic shock & influenza & necrosis & antibiotics & CPR & respiratory failure & CO reduces oxygen transport \\
asthma & TB & bleeding & saline & vaccination & hypotension & ventricular fibrillation \\
pneumonia & hepatitis viruses & lesion & antiviral & bronchoscopy & hypercapnia & significant impedance increases \\
COPD & measles & cell swelling & quercetin & intubation & tachypnea & ventricular fibrillation \\
gastroenteritis & pandemic influenza & hemorrhage & NaCl & transfection & respiratory distress & pulseless electrical activity \\
viral infections & seasonal influenza & diarrhea & ribavirin & bronchoalveolar lavage & hypoxaemia & mildmoderate hypothermia \\
SARS & rabies & toxicity & Norwalk agent & autopsy & pyrexia & cardiogenic oscillations\\
\hline
\end{tabular}
}
\end{table*}

\begin{table}[htb!]
\caption{A sample assertion status labels for a set of entities detected by an NER model as \textit{Disease\_Syndrome\_Disorder} out of CORD-19 dataset.}
\centering
\label{tab:cord_ner_assertions}
\begin{tabular}{lll}
\hline
chunk & assertion \\\hline
systemic disease  & Present \\
skin diseases & Someone-else \\
vascular disorders & Possible \\
infectious diseases & Hypothetical \\
disorders of immunity & Hypothetical \\
infectious disease & Hypothetical \\
word malacia & Present \\
chapter-necrosis & Hypothetical \\
parvovirus & Someone-else \\\hline
\end{tabular}
\end{table}

\section{Benchmarking Speed and Scalability}
\label{sec:scaling}

The design of Spark NLP pipelines as described in Figure~\ref{fig:pipeline_diagram}, where new columns are added to an existing (potentially distributed) data frame with each additional pipeline step, is optimized for parallel execution. It's design for the case where different rows may reside on different machines - benefiting from the optimizations and design of Spark ML.

\begin{figure}[ht!]
\includegraphics[width=0.9\columnwidth, scale=0.9]{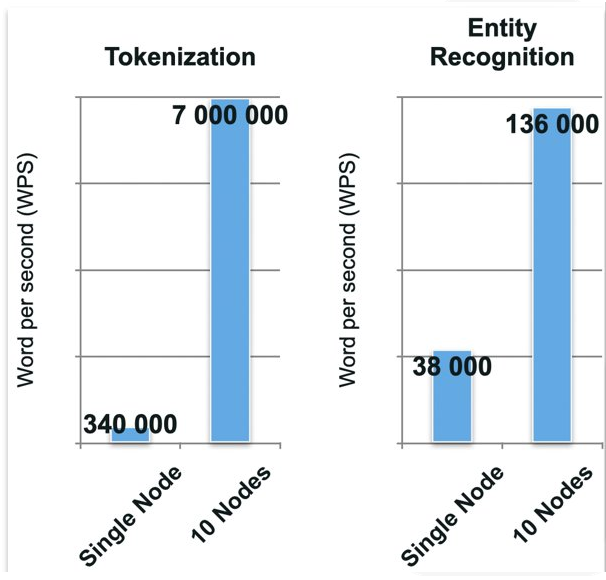}
\centering
\caption{Comparing the Spark NLP document parsing pipeline in standalone and cluster mode. Tests show that tokenization is 20x faster while the entity extraction is 3.5x faster in cluster mode when compared to standalone mode.}
\label{fig:scale_bars}
\end{figure}

In order to evaluate how fast the pipeline works and how effectively it scales to make use of a compute cluster, we ran the same Spark NLP prediction pipelines in local mode and in cluster mode. In local mode, a single Dell server with 32 cores and 32 GB memory was used. In cluster mode, 10 machines with 32 GB and 16 cores each were used, in a Databricks cluster on AWS. The performance results are shared in Figure~\ref{fig:scale_bars}. 

These benchmarks show that tokenization is 20x faster while the entity extraction is 3.5x faster on the cluster, compared to the single machine run. It indicates that speedup depends on the complexity of the task. For example, tokenization provides super-linear speedup (i.e. growing machines by 10x improves speed by more than 10x), while NER delivers sub-linear speedup (because it's a more computationally complex task).

\section{Conclusion}
\label{sec:conclusion}

In this study, we introduced a set of pretrained named entity recognition and assertion status detection models that are trained on biomedical and clinical datasets with deep learning architectures on top of Spark NLP. We then present how to extract relevant facts from the CORD-19 dataset by applying state-of-the-art NER and assertion status models in a unified \& scalable pipeline and shared the results to illustrate extracting valuable information from scientific papers.

The results suggest that papers present in the CORD-19 include a wide variety of the many entity types that this new NLP pipeline can recognize, and that assertion status detection is a useful filter on these entities. This bodes well for the richness of downstream analysis that can be done using this now structured and normalized data - such as clustering, dimensionality reduction, semantic similarity, visualization, or graph-based analysis to identity correlated concepts. One future research direction is to apply these downstream analyses on the richer, scalable, and more accurate insights that this NLP pipeline generates.

Since NER and assertion status models in Spark NLP are trainable, it is easy to add support for a new language like German, French, or Spanish, as long as there is a annotated data for it. Spark NLP currently supports 46 languages and 3 languages for Healthcare - English, German and Spanish. Spark NLP provides production-grade libraries for popular programming languages - Python, Scala, Java and R - and has an active community, frequent releases, public documentation and freely available code examples. Future work in this space includes adding support for additional languages, additional entity types, and extending the NLP pipeline further by adding relation extraction and entity resolution models.

\bibliography{References}

\appendix

\newpage

\section*{Appendices}
\label{sec:appendix}

\section{NER Model Training Tagging Schema}
\label{appendix:tagging}

BIO (Begin, Inside and Outside) and BIOES (Begin, Inside, Outside, End, Single) schemes for encoding entity annotations as token tags. Words tagged with O are outside of named entities and the I-XXX tag is used for words inside a named entity of type XXX. Whenever two entities of type XXX are immediately next to each other, the first word of the second entity will be tagged B-XXX to highlight that it starts another entity. On the other hand, BIOES (also known as BIOLU) is a little bit sophisticated annotation method that distinguishes between the end of a named entity and single entities. BIOES stands for Begin, Inside, Outside, End, Single. In this scheme, for example, a word describing a gene entity is tagged with “B-Gene” if it is at the beginning of the entity, “I-Gene” if it is in the middle of the entity, and “E-Gene” if it is at the end of the entity. Single-word gene entities are tagged with “S-Gene”. All other words not describing entities of interest are tagged as ‘O’.

\section{Defining a Spark NLP Pipeline}
\label{appendix:python_code_pipeline}

\begin{lstlisting}[language=Python]

from sparknlp_jsl.annotator import *

documentAssembler = DocumentAssembler()\
  .setInputCol("text")\
  .setOutputCol("document")

sentenceDetector = SentenceDetector()\
  .setInputCols(["document"])\
  .setOutputCol("sentence")

tokenizer = Tokenizer()\
  .setInputCols(["sentence"])\
  .setOutputCol("token")

word_embeddings = WordEmbeddingsModel.pretrained("embeddings_clinical", "en", "clinical/models")\
  .setInputCols(["sentence", "token"])\
  .setOutputCol("embeddings")

clinical_ner = NerDLModel.pretrained("ner_clinical", "en", "clinical/models") \
  .setInputCols(["sentence", "token", "embeddings"]) \
  .setOutputCol("ner")

ner_converter = NerConverter() \
  .setInputCols(["sentence", "token", "ner"]) \
  .setOutputCol("ner_chunk")

clinical_assertion = AssertionDLModel.pretrained("assertion_dl", "en", "clinical/models") \
    .setInputCols(["sentence", "ner_chunk", "embeddings"]) \
    .setOutputCol("assertion")
    
nlpPipeline = Pipeline(stages=[
    documentAssembler, 
    sentenceDetector,
    tokenizer,
    word_embeddings,
    clinical_ner,
    ner_converter,
    clinical_assertion
    
\end{lstlisting}

\section{Training an NER Model in Spark NLP}
\label{appendix:python_code_ner}

\begin{lstlisting}[language=Python]

from pyspark.ml import Pipeline
import sparknlp
from sparknlp.training import CoNLL
from sparknlp.annotator import *

spark = sparknlp.start()

training_data = CoNLL().readDataset(spark, 'BC5CDR_train.conll')

word_embedder = WordEmbeddings.pretrained('wikiner_6B_300', 'xx') \
 .setInputCols(["sentence",'token'])\
 .setOutputCol("embeddings")

nerTagger = NerDLApproach()\
  .setInputCols(["sentence", "token", "embeddings"])\
  .setLabelColumn("label")\
  .setOutputCol("ner")\
  .setMaxEpochs(10)\
  .setDropout(0.5)\
  .setLr(0.001)\
  .setPo(0.005)\
  .setBatchSize(8)\
  .setValidationSplit(0.2)\

pipeline = Pipeline(
    stages = [
    word_embedder,
    nerTagger
  ])

ner_model = pipeline.fit(training_data)
\end{lstlisting}

\section{Pretrained NER Models and Entities Covered}
\label{appendix:entities}

\subsection{ner\_anatomy\_coarse}

~\cite{pyysalo2014anatomical}

\textbf{Entities:} anatomy    

\subsection{ner\_anatomy}                          

\textbf{Entities:} organism\_substance, organ, cellular\_component, immaterial\_anatomical\_entity, tissue, organism\_subdivision, anatomical\_system, cell, pathological\_formation, developing\_anatomical\_structure, multi                                  

\subsection{ner\_bionlp}

\cite{nedellec2013overview}

\textbf{Entities:} cellular\_component, organ, cancer, organism\_substance, multi, simple\_chemical, tissue, anatomical\_system, organism\_subdivision, immaterial\_anatomical\_entity, organism, developing\_anatomical\_structure, amino\_acid, gene\_or\_gene\_product, pathological\_formation, cell                                                                       

\subsection{ner\_cellular} 

~\cite{kim2004introduction}

\textbf{Entities:} dna, cell\_line, cell\_type, rna, protein                                                                                    

\subsection{ner\_clinical}      

~\cite{uzuner20112010}

\textbf{Entities:} treatment, problem, test                                                                                                    

\subsection{ner\_deid}

~\cite{stubbs2015identifying}

\textbf{Entities:} location, contact, date, profession, name, age, id                                                                           

\subsection{ner\_deid\_enriched}

~\cite{stubbs2015identifying}

\textbf{Entities:} idnum, country, date, profession, medicalrecord, username, organization, zip, id, healthplan, location, device, hospital, city, email, doctor, street, state, patient, bioid, url, phone, fax, age                                                     

\subsection{ner\_diseases}

~\cite{dougan2014ncbi}

\textbf{Entities:} disease                                                                              

\subsection{ner\_drugs}

~\cite{henry20202018}, ~\cite{segura2013semeval}

\textbf{Entities:} drug                                                                                                                        

\subsection{ner\_events\_clinical}

~\cite{sun2013evaluating}

\textbf{Entities:} test, problem, clinical\_dept, occurrence, date, time, evidential, treatment, frequency, duration   

\subsection{jsl\_ner\_wip\_clinical}    

(in-house annotations from mtsamples and MIMIC-III~\cite{johnson2016mimic})

\textbf{Entities:} triglycerides, oncological, female\_reproductive\_status, form, time, date, alcohol, medical\_history\_header, race\_ethnicity, temperature, drug\_brandname, frequency, fetus\_newborn, sexually\_active\_or\_sexual\_orientation, disease\_syndrome\_disorder, section\_header, social\_history\_header, strength, cerebrovascular\_disease, family\_history\_header, employment, weight, pregnancy, total\_cholesterol, diet, ekg\_findings, gender, drug\_ingredient, vaccine, substance, oxygen\_therapy, internal\_organ\_or\_component, blood\_pressure, overweight, obesity, birth\_entity, heart\_disease, diabetes, substance\_quantity, treatment, death\_entity, route, modifier, test, clinical\_dept, communicable\_disease, psychological\_condition, hypertension, direction, o2\_saturation, hyperlipidemia, imagingfindings, vs\_finding, allergen, dosage, kidney\_disease, bmi, smoking, pulse, ldl, symptom, labour\_delivery, relationship\_status, external\_body\_part\_or\_region, hdl, respiration, procedure, height, vital\_signs\_header, relativetime, relativedate, injury\_or\_poisoning, medical\_device, test\_result, duration, age, admission\_discharge, ner\_medmentions\_coarse, pathologic\_function, geographic\_area, group, diagnostic\_procedure, organic\_chemical, organism\_attribute, mental\_or\_behavioral\_dysfunction, organization, research\_activity, therapeutic\_or\_preventive\_procedure, biomedical\_or\_dental\_material, mammal, genetic\_function, body\_system, substance, daily\_or\_recreational\_activity, quantitative\_concept, health\_care\_activity, molecular\_function, indicator,\_reagent,\_or\_diagnostic\_aid, body\_substance, virus, eukaryote, disease\_or\_syndrome, spatial\_concept, anatomical\_structure, body\_part,\_organ,\_or\_organ\_component, laboratory\_procedure, sign\_or\_symptom, nucleic\_acid,\_nucleoside,\_or\_nucleotide, food, mental\_process, prokaryote, nucleotide\_sequence, professional\_or\_occupational\_group, cell, biologic\_function, manufactured\_object, molecular\_biology\_research\_technique, gene\_or\_genome, chemical, neoplastic\_process, pharmacologic\_substance, tissue, qualitative\_concept, amino\_acid,\_peptide,\_or\_protein, fungus, population\_group, body\_location\_or\_region, clinical\_attribute, injury\_or\_poisoning, medical\_device, cell\_component, plant

\subsection{ner\_posology} 

~\cite{henry20202018}

\textbf{Entities:} form, dosage, strength, drug, route, frequency, duration                                                                     

\subsection{ner\_risk\_factors}

~\cite{stubbs2015identifying}

\textbf{Entities:} family\_hist, smoker, obese, medication, hypertension, hyperlipidemia, phi, diabetes, cad                               

\subsection{ner\_human\_phenotype\_go\_clinical}

~\cite{sousa2019silver}

\textbf{Entities:} go, hp                                                                                                                       

\subsection{ner\_human\_phenotype\_gene\_clinical}

~\cite{sousa2019silver}

\textbf{Entities:} gene, hp                                                                                                                     

\subsection{ner\_chemprot\_clinical}

\textbf{Entities:} gene, chemical

\subsection{ner\_ade\_clinical}

\textbf{Entities:} ade, drug                                                                                                                    

\subsection{ner\_chemicals}

\textbf{Entities:} chem                                                                                                                         

\subsection{ner\_bacterial\_species}

\textbf{Entities:} species        

\end{document}